\documentclass[letterpaper, 10 pt, conference]{IEEEtran}  

\pdfoutput=1  
\IEEEoverridecommandlockouts                              

\usepackage{graphics} 
\usepackage{epsfig} 
\usepackage{mathptmx} 
\usepackage{times} 
\usepackage{amsmath} 
\usepackage{amssymb}  
\usepackage[T1]{fontenc}
\usepackage{amsfonts}
\usepackage{booktabs}
\usepackage{siunitx}
\usepackage{caption} 
\usepackage{booktabs}
\usepackage{graphicx}
\usepackage{algorithm}
\usepackage{algorithmic}
\usepackage{float}

\usepackage{caption}
\usepackage{array, caption, tabularx, makecell, booktabs}%

\usepackage{geometry}
\usepackage{comment}
\usepackage[caption=false]{subfig}

\title{\LARGE \bf
Spatial Imagination With Semantic Cognition for Mobile Robots
}

\author{Zhengcheng Shen$^{1}$\thanks{$^{1}$ Zhengcheng Shen, Linh K{\"a}stner and Jens Lambrecht are with the Chair Industry Grade Networks and Clouds, Faculty of Electrical Engineering, and Computer Science,				
		Berlin Institute of Technology, Berlin, Germany
		{\tt\small zhengcheng.shen@campus.tu-berlin.de}}, Linh K{\"a}stner$^{1}$ and Jens Lambrecht$^{1}$
}

\begin{document}

\maketitle
\thispagestyle{empty}
\pagestyle{empty}


\begin{abstract}
The imagination of the surrounding environment based on experience and semantic cognition has great potential to extend the limited observations and provide more information for mapping, collision avoidance, and path planning. This paper provides a training-based algorithm for mobile robots to perform spatial imagination based on semantic cognition and evaluates the proposed method for the mapping task. We utilize a photo-realistic simulation environment, Habitat, for training and evaluation. The trained model is composed of Resent-18 as encoder and Unet as the backbone. We demonstrate that the algorithm can perform imagination for unseen parts of the object universally, by recalling the images and experience and compare our approach with traditional semantic mapping methods. It is found that our approach will improve the efficiency and accuracy of semantic mapping.


\end{abstract}

\section{Introduction}
Aristotle described the imagination as a faculty in humans (and most other animals) that produces, stores, and recalls the images used in a variety of cognitive activities, including those, which motivate and guide action \cite{sep-aristotle-psychology}. The topic about how to mimic the imagination with modern neural network and to benefit the industries is open and intensively researched by a large group of scientists. The Imagination-Augmented Agents in \cite{weber_imagination-augmented_2018} utilizes environment models, which take the present information as input and predict the future to benefit the decision. The novel Reinforcement Learning framework Dream in \cite{hafner_dream_2020} uses a Markov decision process for imagination, which can significantly accelerate the training process. Semantic labels are used by the imagination model in \cite{churamani_clifer_2020} to generate additional data for the enhancement of the facial expression recognition. Imagination is also used for efficient mapping in \cite{ramakrishnan_occupancy_2020,mohajerin_multi_step_2019}.
\\
Human imagination is built on top of our complex cognitive functions. The human brain can encode different cognitive information in modality-specific patterns and all the cross-modal interactions for all modality-specific sources of information are mediated \cite{ralph_neural_2017}. Learning from multiple modalities could be essential to build machine imagination. Besides, the ability to process multi-modal information like visual, semantic and sound information is important for the interaction between robots and humans. The Semantic maps developed in  \cite{mccormac_semanticfusion_2016,chaplot_object_2020, Object_Semantic_Grid_Mapping2020} provide the necessary information for intuitive user interaction and intelligent path planning.
 \begin{figure}[ht]
	\centering
	\includegraphics[width=2.5in]{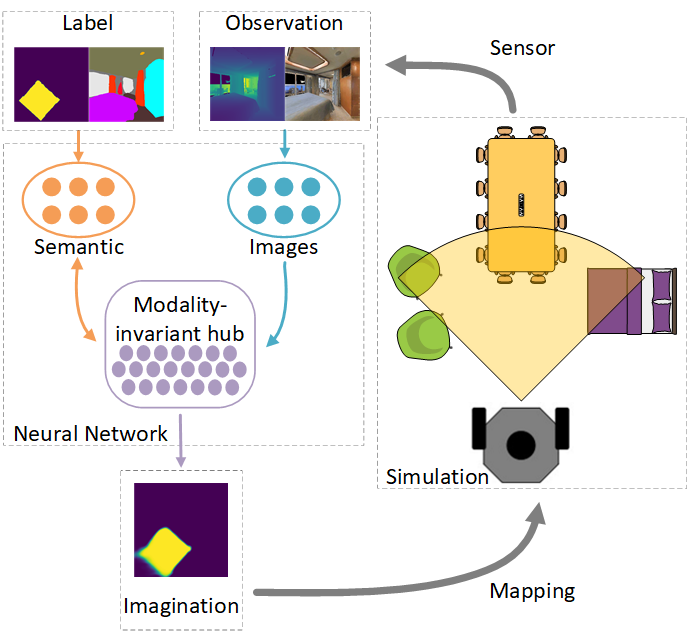}
	\caption{An overview of the concept. The robot inside the simulation environment has a sensor, which provides RBG-D observations. The observation will go through a visual encoder for feature extracting. Semantic labels are used for generating the ground truth data for training. Semantic cognition and imagination will be learned by a modality-invariant hub, which is simulated by an Unet. The output of the neural network is the imagination of the 2D top-down-view of a specific category. As a result, the imagination enhances the robot's understanding of the surrounding environment, which in turn benefits mapping, path planning or other applications}
	\label{fig:intro}
\end{figure}
In this paper, we propose an algorithm that can perform imagination based on observation and semantic cognition. The overview of the concept is shown in Fig. \ref{fig:intro}. A spatial imagination model based on semantic cognition is hard to train without a proper simulation environment. The recently published simulation environment Habitat \cite{habitat19iccv} enables us to realize the implementation and training process. The objective of the imagination model is to draw a multiple-layer-2D-semantic-map. Furthermore, a mapping task with a set of unseen scenes is implemented for evaluation purposes.
The main contributions of this work are the following:
\begin{itemize}
    \item A mapping algorithm for 2D semantic maps equipped with imagination ability. The improved efficiency of the mapper based on imagination enables the robot to explore the new environment faster using prior knowledge. Additionally, imagination can offer the robot further information about the surrounding environment for decision making.
    \item Comparison of our model with a normal mapper based on a semantic-segmentation-neural-network. A significant margin on efficiency and accuracy is shown between the proposed method and the widely used 2D semantic mapping method in \cite{chaplot_object_2020,Object_Semantic_Grid_Mapping2020}.
\end{itemize}
The paper is structured as follows. Sec. II begins with related works followed by our problem statement. Subsequently, the methodology is presented in Sec III. The details about the implementation are described in Sec. IV. Sec. V presents the results and discussion. Finally, Sec. VI will give a conclusion and outlook.

\newgeometry{top=0.75in,bottom=0.75in,right=0.75in,left=0.75in}
\section{Related Works}
\textit{Semantic mapping} A two dimensional grid map is widely used in mobile robots for global path planning and local path planning \cite{daniel_theta_2010,Lavalle98rapidly_exploringrandom}. For a long time, it records only the information for occupancy objects and serves well for simple navigation purposes. However, the extension of rich semantic information within a grid map opens a door for further functionalities. In \cite{Semantic_Mapping_Spatial_Concepts_2019}, a semantic map with a spatial concept was built according to visual input and word information. In \cite{Object_Semantic_Grid_Mapping2020},  LiDAR and RGB-D sensors are used for semantic mapping. The authors generated a semantic segmentation image from the RGB-D sensor and turned it into a segmented point cloud based on LiDAR sensor. The top-down-view map is then generated from the segmented point cloud. However, both works are limited by the number of scenes and haven't tested the model in a data set with a large group of variations. In this research, Habitat is used for training and evaluation, which enables us to use 
all scenes from Matterport3D dataset for training and evaluation \cite{Matterport3D}. The same simulation environment was used in \cite{chaplot_object_2020} for object goal navigation. Within the research, a multiple-layer semantic map is generated by a Mask-RCNN \cite{he_mask_2018}, which is pretrained with MS-COCO \cite{MS-coco}, and additional depth information. The same framework is also used in \cite{Object_Semantic_Grid_Mapping2020}. One drawback for the approach is that a slight error in the first-person segmentation model will cause a high amount of noise to the map after projection. A denoising neural network is deployed in \cite{chaplot_object_2020} to mitigate the error, while a special algorithm is used in \cite{Object_Semantic_Grid_Mapping2020}. Nevertheless, the process will reduce the information contained in the map again. The most significant improvement of our model can anticipate the unseen area based on former experience and semantic cognition, which will improve the accuracy and efficiency of semantic mapping.
\\
\textit{Scene Completion} Although the inferring unseen 3D structure and semantics are intensively researched in SSCNet \cite{song2016semantic}, it is computationally expensive to use 3D map information for a mobile robot, and the field of view (FoV) of the observation is not extended. The recently published research \cite{ramakrishnan_occupancy_2020} anticipates the unseen area with limited observations though it hasn't introduced any semantic information into the system manually. Furthermore, the area for imagination is also not well limited. The imaginable area must have correlation with the observation. Otherwise, it will introduce additional noise to the model. In this paper, we will follow a similar approach for imagination but with semantic information alongside. 
\\
\section{Conceptual Design}
\subsection{Problem Statement}
The ability to adapt to a new environment is a important measurement for an intelligent robot. It highly depends on the way to utilize the former experience and the information that it records during the exploration. To balance between the complexity of computation and the information of the world, a multi-layer semantic map \cite{chaplot_object_2020} is designed. The widely used method, which generates a voxel representation from the first person RGB-D images, can only classify the pixel by previous experience but not fully utilize the rich features inside the observation to perform imagination. Besides, imagination can help the agent to obtain a better understanding of the surrounding environment,  which may lead to a better decision. All these requirements for an intelligent robot results in the open questions, how can we enable our robot to imagine and how can imagination benefit the industries?

\subsection{System Design}
\begin{figure*}[ht]
	\centering
	\includegraphics[clip, width=5.9in,trim={1.cm 4.0cm 1cm 2.5cm}]{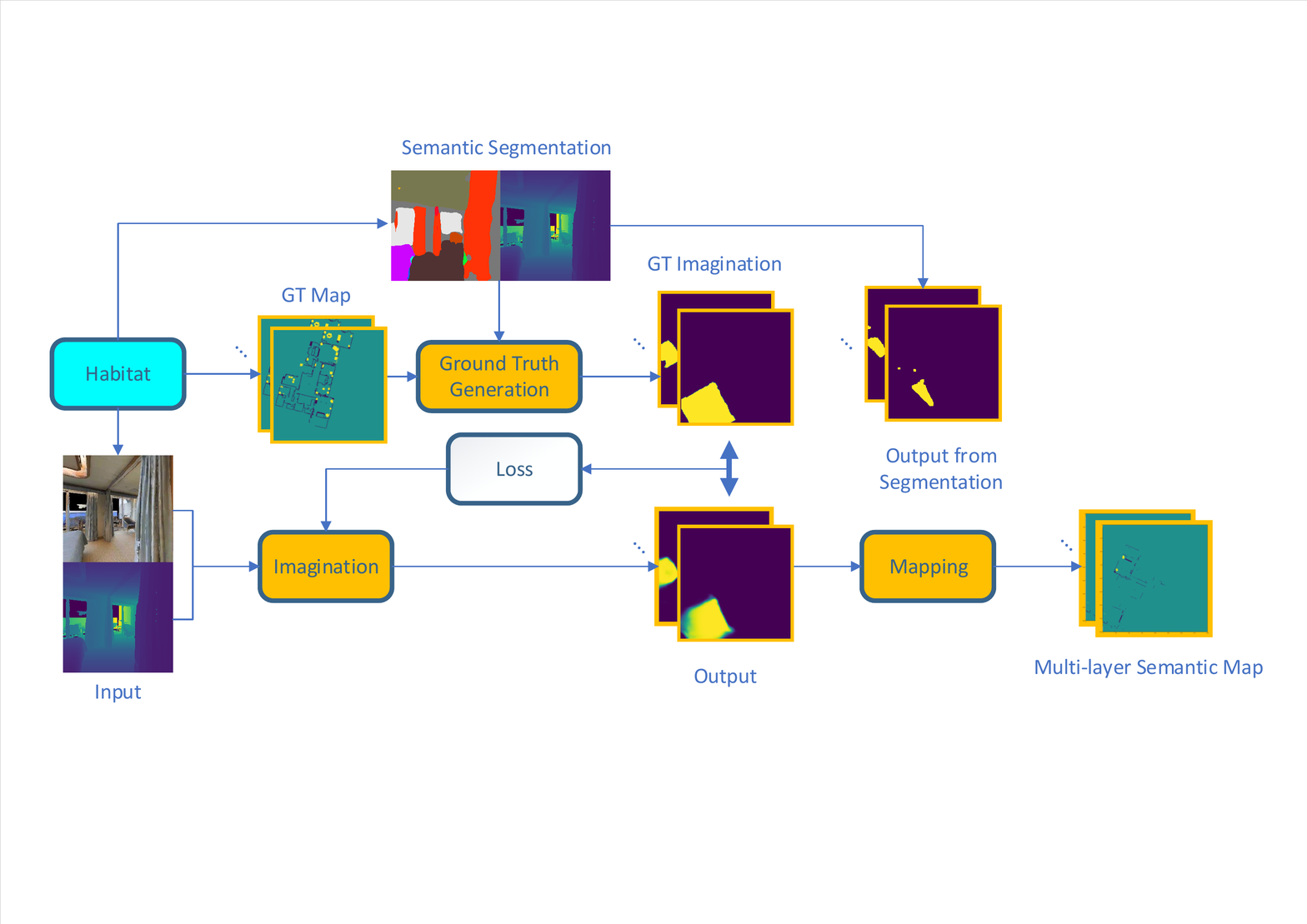}
	\caption{System design of the mapping process. The input of the imagination module is RBG-D data and the output should be a top-down view of the current observation area. The ground truth is generated based on a high-resolution map and semantic information. It does not only extend the FoV of the robot but also inpaints the unseen occluded regions. The mapping module registers the observation into the global map with accurate coordinates from the simulator.}
	\label{concept}
\end{figure*}

The overall framework for semantic mapping is described in Fig. \ref{concept}. The process contains three main functions, namely the functions for ground truth generation, imagination, and mapping. The ground truth map is first generated inside the 3D simulator, Habitat, with a high sample resolution (2 cm between two consecutive pixels). Combined with the semantic information provided by manually labeled scenes, the ground truth is generated for the expected imagination. The imagination function is responsible for both object detection and imagination since the ground truth only provides the imagination for a certain kind of object. Furthermore, it also contains a denoise module for cleaning the overmuch imagination. During the training process, the agent will learn the regulation of a human-designed object and its semantic concept. After training, it will obtain the ability to perform the imagination on new objects that haven't been seen by the agent before. The method does not only extend the FoV of the robot but also inpaints the unseen occluded regions of the object in a 2D manner. After the imagination module, a mapping module will register the ego map into the full map by the methods provided in \cite{chaplot2020learning} with known position from the simulator. The map generated from imagination and mapping has multiple layers and each layer is corresponding to a specific object. In the next section, we will detail the implementation of the three different functions.

\section{Implementation}
\subsection{Ground Truth Generation}
The quality of the ground truth information directly influences the performance of the trained model. Moreover, it could influence the training efficiency and decide if the training will converge to a local minimum. The ground truth generation contains two steps: the first step is to generate the ground truth egocentric map for a single class. Subsequently, a mask is generated to filter out the objects that cannot be seen by the agent.
\subsubsection{Object Ground Truth Map}
\begin{figure}[!htp]
    \centering
	\includegraphics[width=0.45 \textwidth]{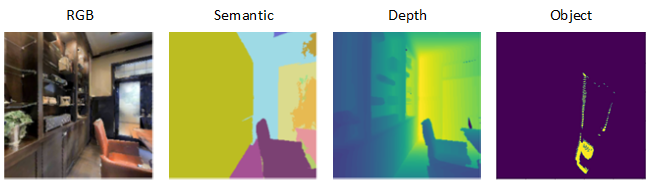}
	\caption{The first three pictures are the sensor's output from Habitat and the last one is the top-down view of the chair generated from the depth image.}
	\label{fig:noise from segementation}
\end{figure}

Although the semantic label provided by Matterport3D is professionally labeled, it still has numeric errors when the simulator presents the first-person view. The slight error in the segmentation of the first-person view will introduce unignorable noise into the semantic top-down view, which is shown in Fig. \ref{fig:noise from segementation}.
Since Habitat also provides a semantic bounding box for objects, we choose a different strategy to generate the ground truth map. First, we generate a high-resolution occupancy map, $M_{occu}$, in which the real distance between two pixels is 2 cm. The shape of the total map used in training is (2001, 2001). Some scenes with extra large layouts will be discarded during the selection. To generate the normal occupancy map, we sample the points at a distance of 50 cm and register the egocentric occupancy maps for the 360$^{\circ}$  around the points into the global map. At the same time, we also register all the semantic bounding boxes for the specific object into another map, which is marked as $M_{lable}$. We facilitate the ground truth map generated from first-person view with a dilation function and mark the map as $M_{seg}$. The kernel of the dilation function is 30x30. In the end, the final semantic ground truth map will be the intersection of these three:
\begin{equation}
    M_{gt} = M_{occu} \cap M_{lable} \cap M_{seg} \\
\end{equation}
\subsubsection{Imaginable Filter}
An imaginable filter indicates weather the cells are imaginable. The egocentric semantic map directly that cuts from the ground truth map may include the objects behind a wall, which is barely impossible to imagine according to the observations. Unimaginable objects will make the system more inconsistent. Thus, we filter them out of the ground truth. The filter used in \cite{ramakrishnan_occupancy_2020} works by simply drawing lines from the agent's current location and stopping once a wall is hit. It cannot be used for our use case since it avoids the imagination about an object, which is partially behind the wall. During the training process, we will obtain an egocentric semantic top-down view, $V_{object}$ and an egocentric seen area top-down,$V_{seen}$.
In this case, we propose the imaginable mask as:
\begin{equation}
    V_{filter} = dilation(V_{object}*V_{seen}, \delta)
\end{equation}
where $\delta$ is the kernel matrix for dilation.
\noindent
The size of the dilation kernel can control the size of the imaginable area, which is 50 in our experiment.

\subsection{Imagination}
\begin{figure}[!htp]
    \centering
	\includegraphics[width=0.45 \textwidth]{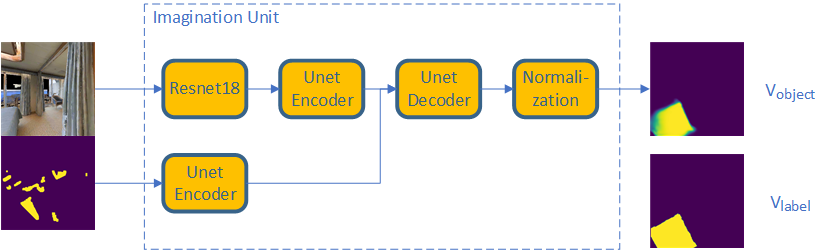}
	\caption{Detailed structure of the imagination unit}
	\label{fig:detail of the imagination unit}
\end{figure}

A suitable neural network framework is another essential part to make the imagination work. The experimental implementation tries to detect three different objects, namely, chair, table and bed. Therefore, we use three identical neural network frameworks, called imagination units, for each category to simplify and accelerate the training speed.
\begin{figure}[!htp]
    \centering
	\includegraphics[width=0.45 \textwidth]{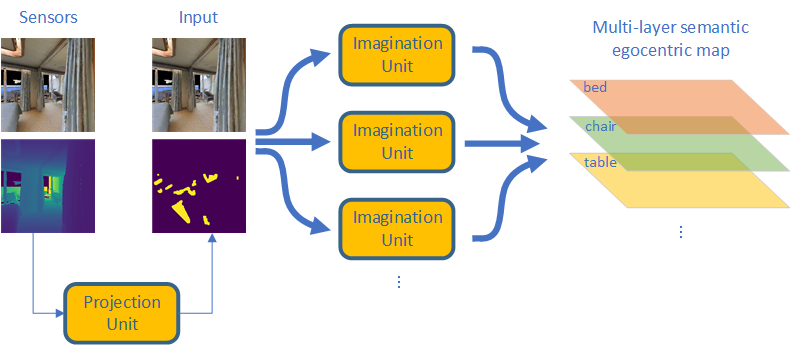}
	\caption{Overview of the imagination module. Each imagination inside the framework is responsible for one categories.}
	\label{fig:overview of imagination module}
\end{figure}
More details about the imagination unit shows in Fig. \ref{fig:detail of the imagination unit}.
Each imagination unit contains a Resnet18 \cite{he2015deep} as visual encoder to extract the features and an Unet as backbone to merge the information and produce the imagination result. At the end, a sigmoid activation function is used for normalization. The structure is used in \cite{ramakrishnan_occupancy_2020} for occupancy anticipation. Unet was originally developed for biomedical image segmentation \cite{ronneberger2015unet}. Therefore, it has the potential to realize the segmentation and space imagination at the same time. One important aspect of the neural-network-based model is the loss function. Due to the sparsity of the object, the ratio between occupied and empty cells is small and varies from each object. To solve the problem, we defined a weighted loss function with two weight matrices. The first weight matrix $W_{\alpha}$ is the weight for the real occupied area for the specific object:
\begin{align}
   W_{\alpha_{ij}} = 
   \begin{cases} 
        w_{\alpha} - 1 &  \mbox{if  } V_{label_{ij}} > 0     \\
        0 &  \mbox{if }   V_{label_{ij}}==0   \\
    \end{cases}  
\end{align}
where,
\begin{equation}
    w_{\alpha} =  Min(\frac{\textit{Number of total cells}}{\textit{Number of cells occupied by object }+ 1}, w_{\alpha_{max}})
\end{equation}
The ground truth of the occupied cell is important, but the imagination can happen around all the seen area:
\begin{equation}
    M_{seen}^* = dilation(V_{object}*V_{seen}, \epsilon)
\end{equation}
 Therefore the weight has to balance the cells again according to:
\begin{align}
   W_{\gamma_{ij}} = 
   \begin{cases} 
        w_{\gamma} - 1 &  \mbox{if  }  M_{seen_{ij}}^* > 0 \mbox{ and }    V_{label_{ij}}==0  \\
        0 &  \mbox{if }    Others  \\
    \end{cases}  
\end{align}

\begin{equation}
    w_{\gamma} = Min(\frac{\textit{Number of total cells}}{\textit{Number of occupied cells} + 1}, w_{\gamma_{max}})
\end{equation}
In the training process, $w_{\alpha_{max}} = 30$, $w_{\gamma_{max}} = 10$ and $\epsilon = 31$. Combined these two weight matrix, the finally weight matrix is:
\begin{equation}
    W = W_{\alpha} +  W_{\gamma} + 1
\end{equation}
And the loss function will be:
\begin{equation}
    L=\sum_i^{V^2} -W_{ij}( V_{label_{ij}}log  V_{object_{ij}} + (1- V_{label_{ij}})log(1-V_{object_{ij}}))
\end{equation}
During the training, a large batch size is more beneficial to obtain a stable weight matrix, which is helpful for convergence. The batch size in our experiment is 108 and the observations are randomly selected in 16 different scenes at the same time.
\subsection{Mapping}
Although imagination can provide more information about the unseen area, it can also introduce extra noise. Thus, false imaginations cannot be avoided. To mitigate this issue, we utilize a post processing module with more regulation, which improves the result significantly. Once again, we apply $M_{seen}^*$ as the filter, which means only the imagination around the seen area is valid. Subsequently, the egocentric observation,$V_{valid}$ is registered into the full map for evaluation purpose.
\begin{equation}
    V_{valid} = V_{object}* M_{seen}^*
\end{equation}

\section{Results and Evaluation}
\begin{figure}[ht]
     \subfloat[The view points generated with cell width of 3 meters \label{fig:users_track}]{%
       \includegraphics[width=0.13\textwidth]{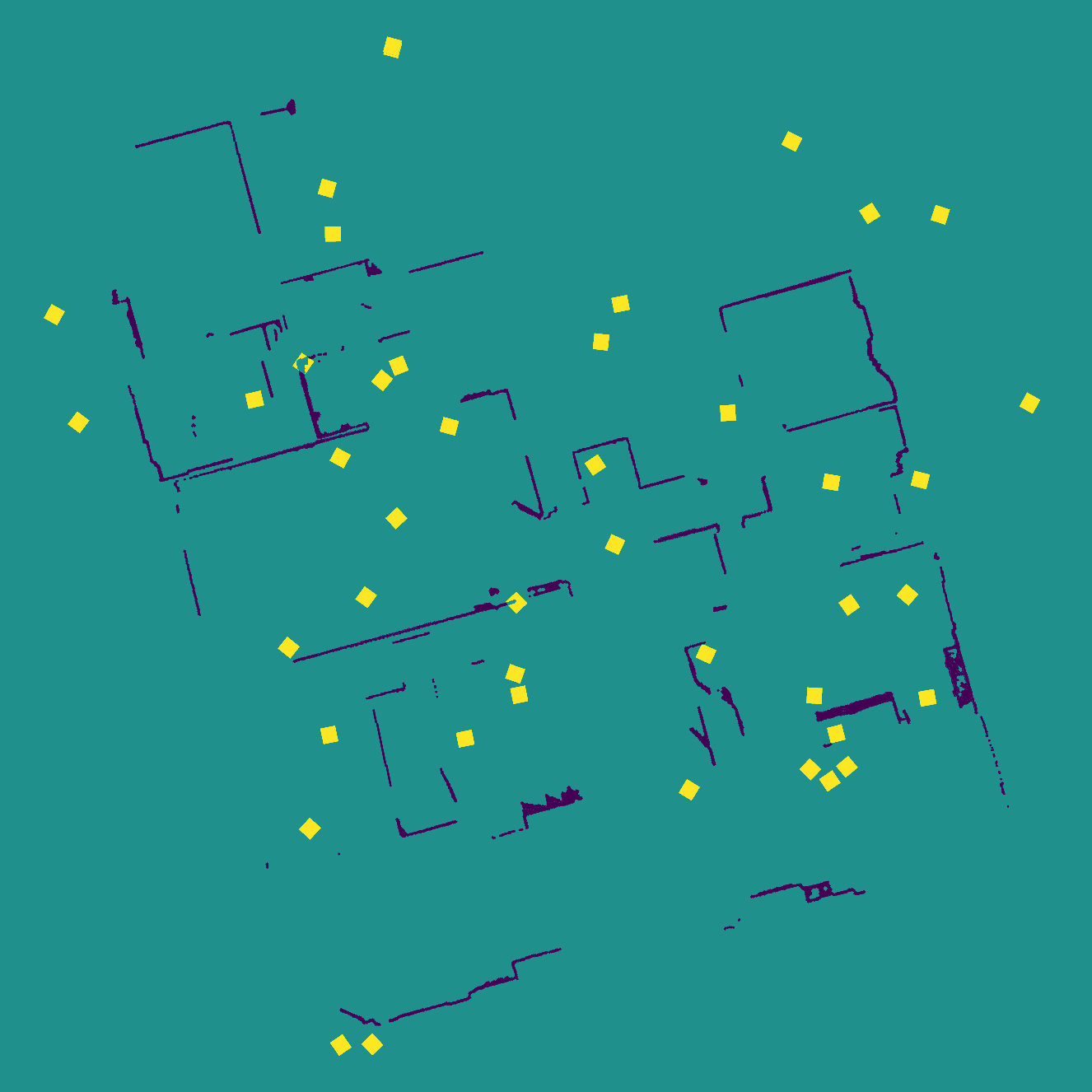}
     }
     \hfill
     \subfloat[Another set of view points generated with cell width of 3 meters \label{fig:Another set of view points generated with cell width of 3 meters}]{%
       \includegraphics[width=0.13\textwidth]{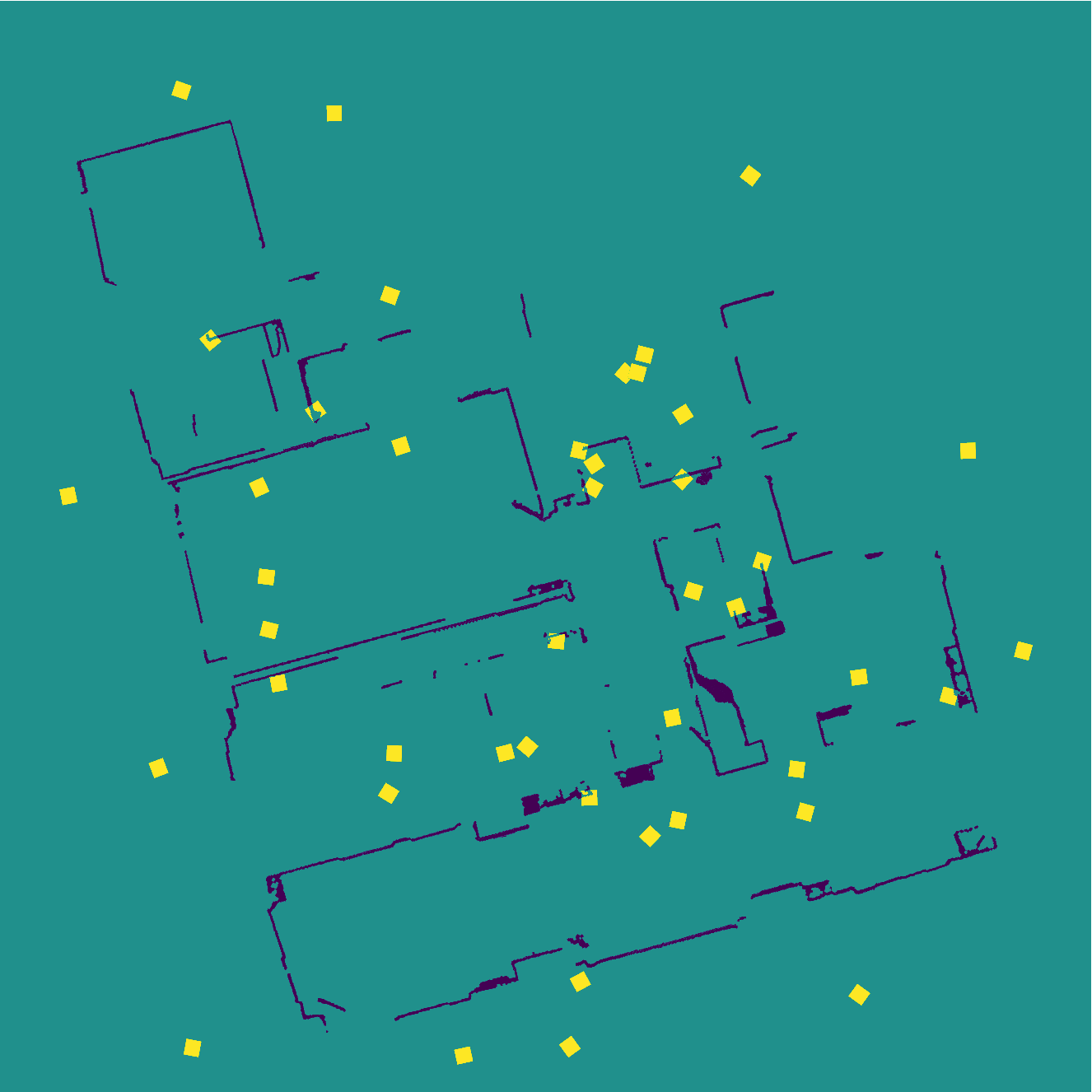}
     }
     \hfill
     \subfloat[The view points generated with cell width of 2 meters \label{fig:The view points generated with cell width of 2 meters}]{%
       \includegraphics[width=0.13\textwidth]{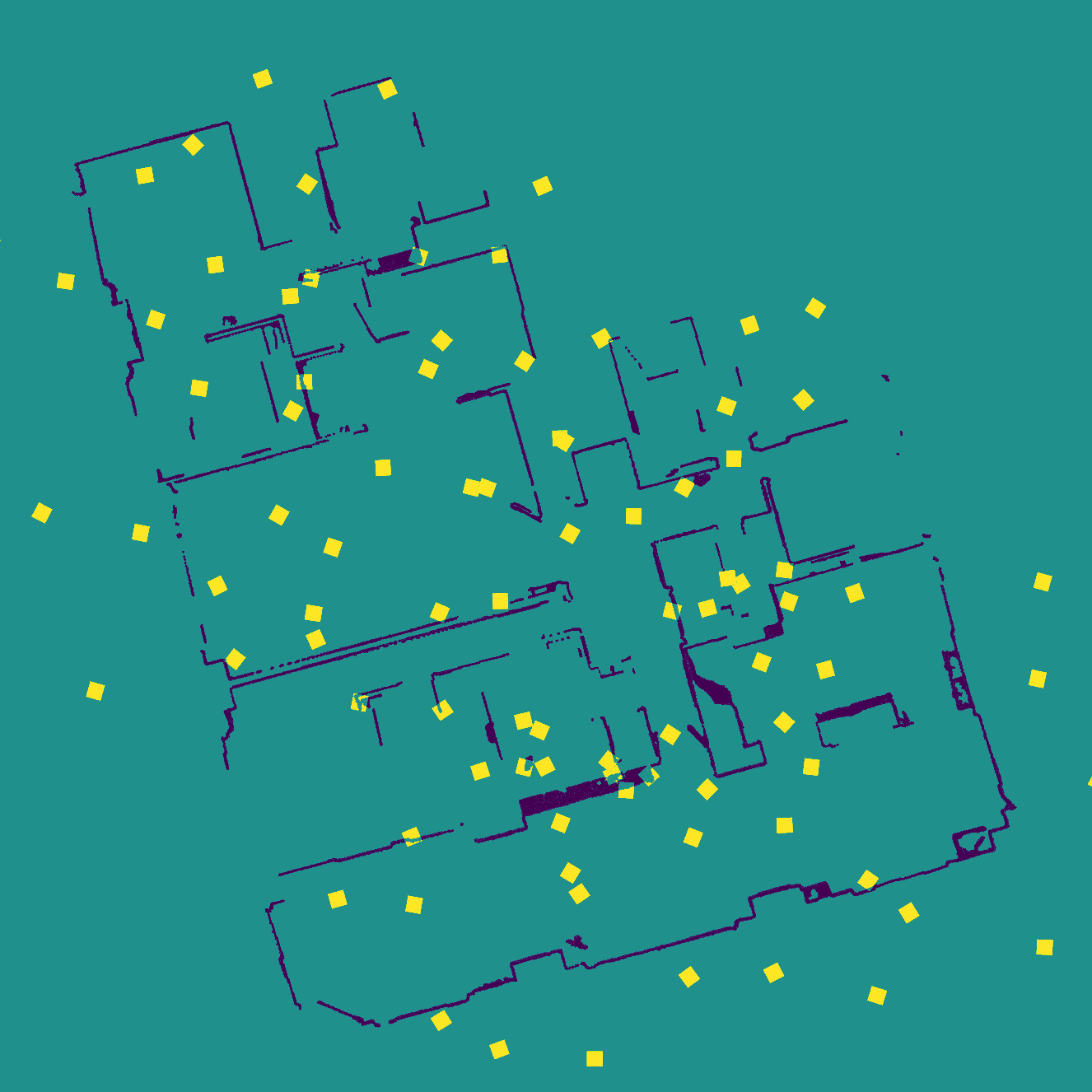}
     }
     \caption{Random generated sparse viewpoints with Algorithm \ref{alg:Sparse View Points}}
     \label{fig:random-selected-view-points}
\end{figure}

To evaluate the proposed approach, we designed the following experiment for a benchmark. For each object, we first prepare 60 scenes, which are different from the training data. This means the robot hasn't seen the environment and style of the objects inside it before. We use the following algorithm \ref{alg:Sparse View Points} to generate a set of sparse points for a map and then find the nearest valid viewpoints based on them.
\begin{algorithm}[!h]
	\caption{Algorithm for Generating Sparse Viewpoints}
	\label{alg:Sparse View Points}
	\begin{algorithmic}
	    \STATE $cell\_width \leftarrow \mbox{preset cell width}$
	    \STATE $coor \leftarrow \mbox{meshrgid with distance of }cell\_width$
		\STATE $M \leftarrow \mbox{get the top down map of the scene}$
		\STATE $offset\_of\_set_x \leftarrow \mbox{uniform} (-\frac{cell\_width}{2}, \frac{cell\_width}{2}) $ 
		\STATE $offset\_of\_set_y \leftarrow \mbox{uniform} (-\frac{cell\_width}{2}, \frac{cell\_width}{2}) $ 
		\FOR{\mbox{each piont }$(p_x,p_y)$ \mbox{in }$coor$}
		
		\IF{$p$ \mbox{has valid value in }$M$}
		\STATE $offset_x \leftarrow \mbox{uniform} (-\frac{cell\_width}{6}, \frac{cell\_width}{6}) $
		\STATE $offset_y \leftarrow \mbox{uniform} (-\frac{cell\_width}{6}, \frac{cell\_width}{6}) $
		\STATE $p_x \leftarrow p_x + offset\_of\_set_x + offset_x $
		\STATE $p_y \leftarrow p_y + offset\_of\_set_y + offset_y $
		\STATE \mbox{add }$(p_x,p_y)$ \mbox{into }$valid\_coor$
		\ENDIF
		\ENDFOR
	\RETURN $valid\_coor$
	\end{algorithmic}
\end{algorithm}
For each view point, the robot will take two observations in two random orientations and try to recover the semantic information for the entire scenes with all the collected information. For each scene, ten different sets of view points will be randomly generated. Three samples of the sparse points are listed in Fig. \ref{fig:random-selected-view-points}.
\begin{figure*}[ht]
	\centering
	\includegraphics[width=4.5in]{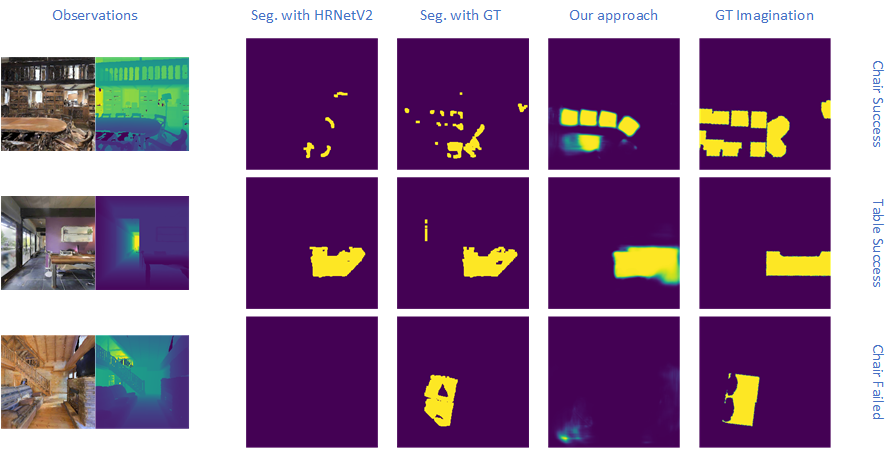}
	\caption{Overall result of a single prediction. The first row shows a successful case of the prediction on the class of chair. The second row shows another successful prediction on the class of table beyond the scope. The last row shows a detection failure on the class of chair.}
	\label{fig:result1}
\end{figure*}
\begin{figure*}[ht]
	\centering
	\includegraphics[width=4.9in]{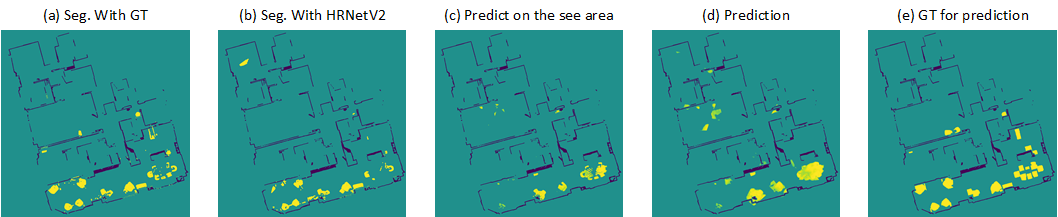}
	\caption{Overall result of the entire map of chairs with cell width of 2 meters.(a) Map comes from segmentation with ground truth. (b) Map comes from segmentation with HRNetV2. (c) Map comes from the seen part of the imagination. (d) Map comes from the entire imagination. (e) Map comes from the ground truth.}
	\label{fig:result2}
\end{figure*}

To quantify the accuracy of a predicted semantic map, we compare the predicted maps with the ground truth of the prediction and calculate the Intersection over Union (IoU). The compared baseline uses the approach described in \cite{chaplot_object_2020} but without the denoising neural network. The Mask-RCNN layer is also replaced with the state-of-art segmentation algorithm HRNetV2 \cite{WangSCJDZLMTWLX19} and the ground truth generated by Habitat. Fig. \ref{fig:result1} shows examples of the predictions with different approaches. Inside the first row, it is evident that our approach successfully predicts the unseen area of chairs. The neural network identified the chairs and imagined the occupied space of the chairs based on the learned experience. The imagination obviously enriches the occupancy prediction. Nevertheless, our approach successfully detects the angle of the chairs and completes the occupied cell according to its imagination. The second row shows an example of the table. The method predicts the unseen part of the desk with a blurred mask. During this imagination, a clean edge cannot be given due to the lack of information. The last row shows a detection failure: the method cannot detect all kinds of chairs due to the limited training scope. In Fig. \ref{fig:result2}, the generated global maps from different approaches are illustrated. It is notable that our method has a lower detection ability compared with the methods with pretrained segmentation neural networks, but the notation area is larger once the object is detected. Besides, the imagination also introduces noise due to false detection, which can be seen in the center part of the map.
\begin{table*}[ht]
\centering\caption {IoU of the Global Map for Selected Approaches} \label{table:IoU}

\begin{tabular}{||p{2.5cm} | p{1 cm}  p{1cm} p{1cm} | p{1 cm}  p{1cm} p{1cm}| p{1 cm}  p{1cm} p{1cm}||}
\toprule

&\multicolumn{9}{c||}{IoU} 
\\
 \hline
Cell width &\multicolumn{3}{c|}{With cell width of 3} & \multicolumn{3}{c|}{With cell width of 2.5}   & \multicolumn{3}{c||}{With cell width of 2}  
\\
 \hline
Category & chair & bed & table & chair & bed & table & chair & bed & table 
\\
 \hline
Seg. with GT & 0.423 & 0.512 & 0.486 & 0.465 & 0.550 & 0.506 & 0.495 & 0.628 & 0.527
\\
 \hline

Seg. with HRNetV2 & 0.155 & 0.379 & 0.189 & 0.183 & \textbf{0.422} & 0.194 & 0.210 & \textbf{0.483} & \textbf{0.214}
\\
\hline
Imagination(seen) & 0.152 & 0.316 & 0.195 & 0.179 & 0.371 & 0.197 & 0.220 & 0.425 & 0.207
\\
 \hline
Imagination & \textbf{0.197} & \textbf{0.386} & \textbf{0.202} & \textbf{0.213} & 0.406 & \textbf{0.205} & \textbf{0.248} & 0.439 & 0.206
\\
\hline
\toprule
\end{tabular}
\end{table*}
\begin{table*}[ht]

\centering\caption {Number of Correct Pixels of the Global Map for Selected Approaches} \label{table:Number of Correct Pixels}

\begin{tabular}{||p{2.5cm} | p{1 cm}  p{1cm} p{1cm} | p{1 cm}  p{1cm} p{1cm}| p{1 cm}  p{1cm} p{1cm}||}
\toprule

&\multicolumn{9}{c||}{Number of Correctly Predicted Pixels} 
\\
 \hline
Cell width &\multicolumn{3}{c|}{With cell width of 3} & \multicolumn{3}{c|}{With cell width of 2.5} & \multicolumn{3}{c||}{With cell width of 2}  
\\
 \hline
Category & chair & bed & table & chair & bed & table & chair & bed & table
\\
 \hline
Seg. with GT & 4420871 & 2229781 & 1880935 & 5816061 & 2717801 & 2373934 & 7169134 & 4506825 & 2648728
\\
 \hline

Seg. with HRNetV2 & 1582407 & 1785445 & 798123 & 2221629 & 2278674 & 1048716 & 2990796 & 3824152 & 1297157
\\
\hline
Imagination(seen) & 1606542 & 1477507 & 863654 & 2256515 & 1994317 & 1103031 & 3288038 & 3294188 & 1344699
\\
 \hline
Imagination & \textbf{2488550} & \textbf{2248037} & \textbf{1246088} & \textbf{3347148} & \textbf{2859842} & \textbf{1617987} & \textbf{4783688} & \textbf{4539208} & \textbf{1884798}
\\
\hline
\toprule
\end{tabular}
\end{table*}
\begin{figure*}[!ht]
     \subfloat[IoUs with chair \label{fig:IoUs with chia}]{%
       \includegraphics[width=0.31\textwidth]{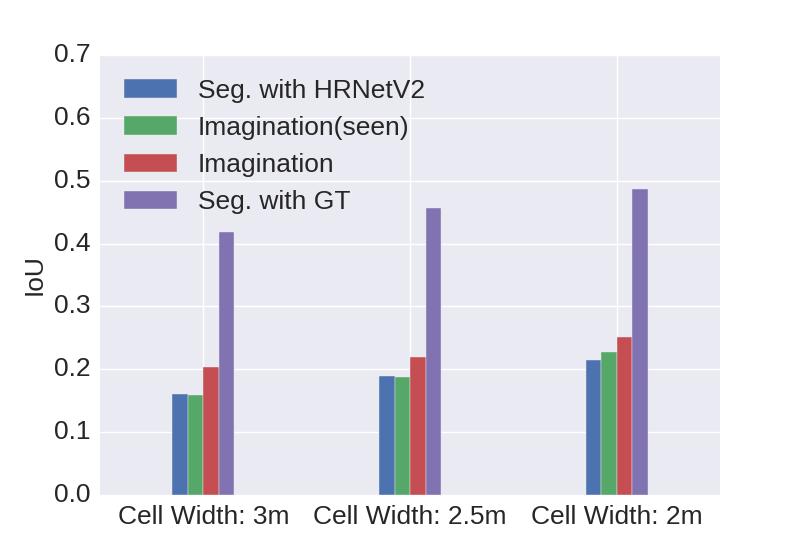}
     }
     \hfill
     \subfloat[IoUs with bed \label{fig:IoUs with bed}]{%
       \includegraphics[width=0.31\textwidth]{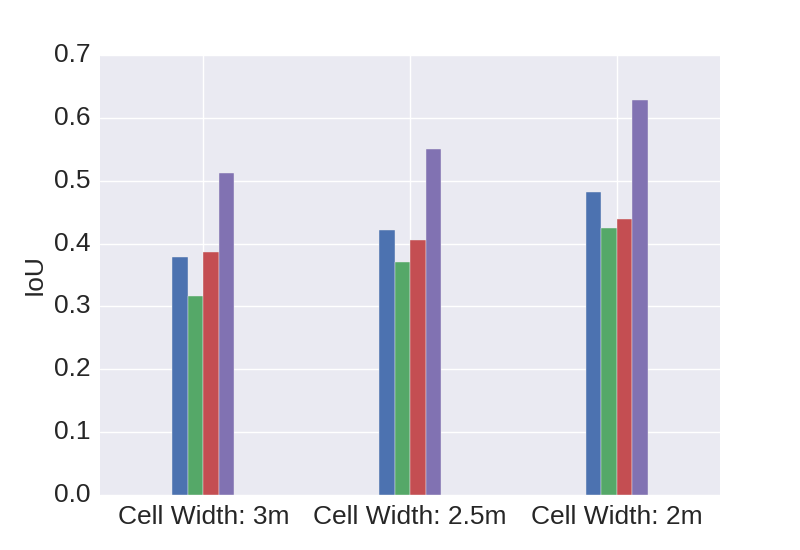}
     }
     \hfill
     \subfloat[IoUs with table \label{fig:IoUs with table}]{%
       \includegraphics[width=0.31\textwidth]{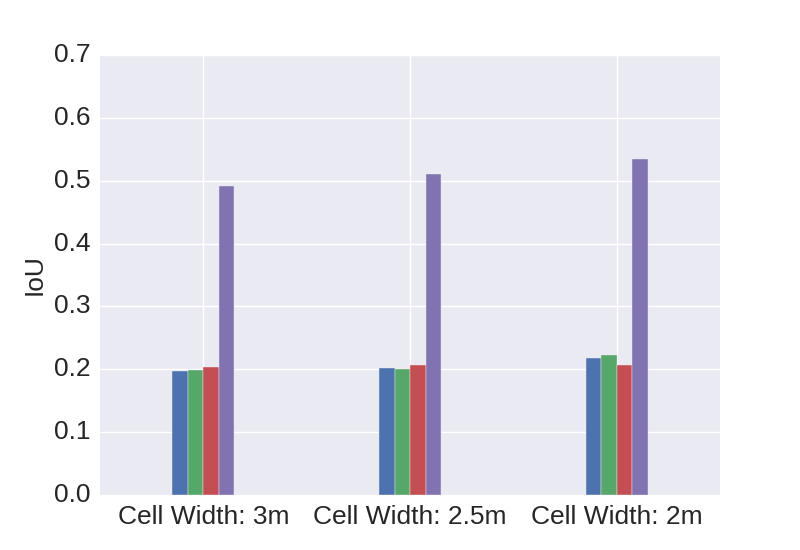}
     }
     \\
    \subfloat[Number of correctly predicted pixels with chair \label{fig:Number of correctly predicted pixels with chiar}]{%
       \includegraphics[width=0.31\textwidth]{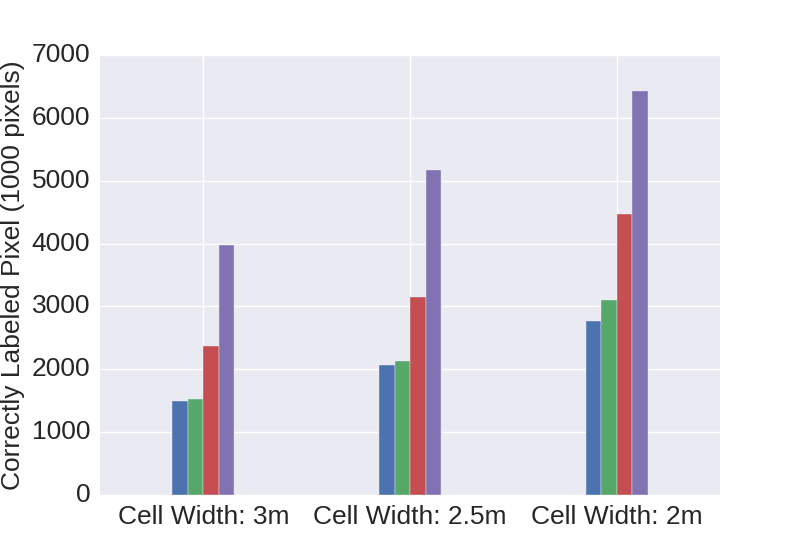}
     }
     \hfill
     \subfloat[Number of correctly predicted pixels with bed \label{fig:Number of correctly predicted pixels with bed}]{%
       \includegraphics[width=0.31\textwidth]{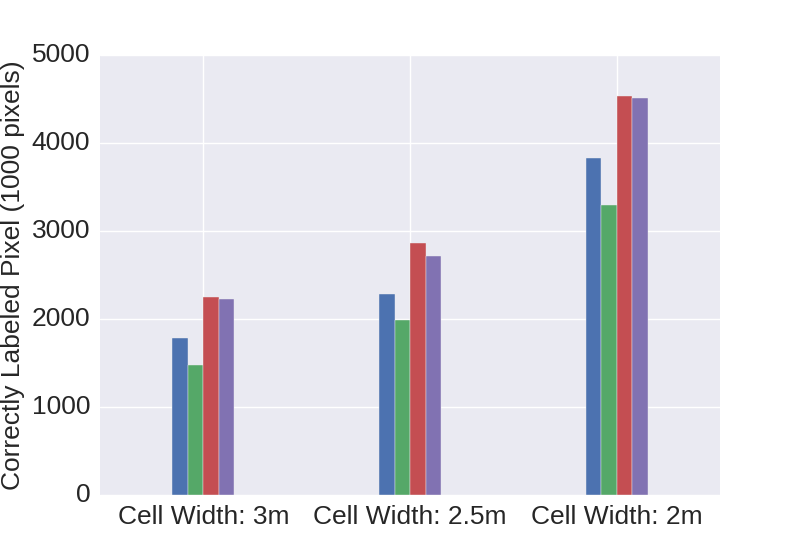}
     }
     \hfill
     \subfloat[Number of correctly predicted pixels with table \label{fig:Number of correctly predicted pixels with table}]{%
       \includegraphics[width=0.31\textwidth]{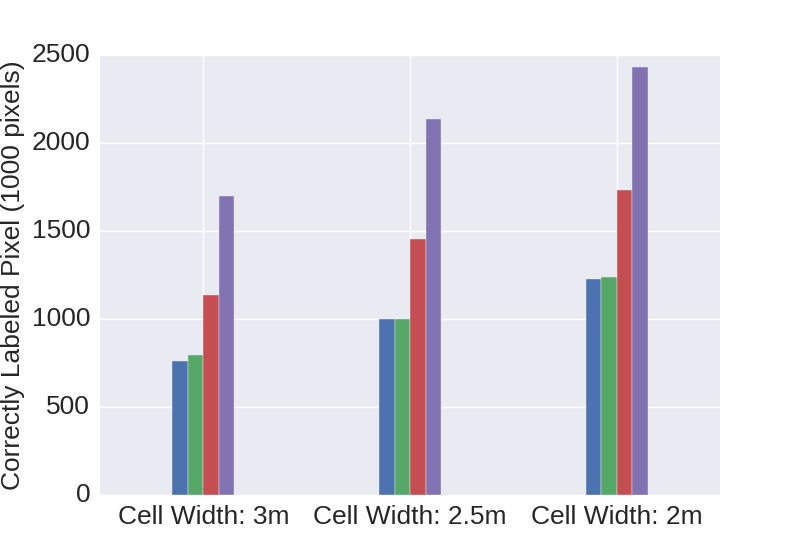}
     }
     \caption{IoUs and the numbers of correctly predicted pixels}
     \label{fig:result3}
\end{figure*}
\\
In Table \ref{table:IoU} and the first row of Fig. \ref{fig:result3} the IoUs of the generated map for different objects with different cell widths are presented. Our approach shows an obvious advantage in regards to the class of chairs when the cell width is 3 meters, by which the viewpoints are more sparse. The reason is that most of the chair are covered by tables and hard to detect by regular methods. Nevertheless, the proposed method shows margin over the regular methods equipped with HRNetV2 on the rest two categories. With the increase of the viewpoints, the margin between the approach and the regular approach becomes smaller, since the space left for imagination shrinks. Compared to the overall data between the seen area of imagination and full imagination, we can conclude that imagination on the unseen area is the main reason for the boost of performance. However, n noticeable advantage compared with HRNetV2 can be observed due to the limited training scope on the styles of the different objects. 
\\
Table \ref{table:Number of Correct Pixels} and the second row of Fig. \ref{fig:result3} show the number of the correctly predicted pixels over the object on the global map. It can be observed that our approach has a large lead due to active imagination on the unseen area with a relatively high IoU. The detected pixels for the bed object of our approach are higher than the regular tool equipped with the ground truth segmentation information. Although the benefits of the imagination decrease with increasing viewpoints, there is still a wide gap between our approach and the regular approaches. We notice that our approach has more detected pixels but not a high IoU. This is caused by false prediction introduced by the active imagination module, which will decrease the performance. Additionally, the detection of an object is not perfect, which means a table could be wrongly classified as a bed.

\section{CONCLUSION}
In this paper, we proposed a method, which can perform the imagination of 2D top-down-view on certain objects based on historical experience and semantic cognition. Furthermore, we show a use case in terms of semantic mapping with the proposed method and evaluate the performance based on this. We combine the imagination with the semantic concept and promote the model to learn the general design of the semantic concept. Afterwards, the learned experience could be utilized to detect an object and perform imagination on the unseen area caused by limited FoV or occlusion. One important contribution of this work is to define the imaginable area, which is the bridge between imagination and observation and makes the training feasible. The method exploits more information from the limited observation, which could bring benefits to different applications. During the semantic mapping, our method shows a substantial performance boost compared with the regular approach with the first-person-view segmentation algorithm. However, it is not the limit of the proposed approach. Our ongoing work focuses on the optimization of the imaginable area, the neural network framework, and the post-processing steps. Furthermore, we we aspire to use a larger data set with better texture and label quality, which could improve the model's performance.

\section*{Appendix}



\begin{table}[H]
\centering
	\setlength{\tabcolsep}{2pt}
	\renewcommand{\arraystretch}{1.2}
		\caption{Hyperparameters for Training}
	\begin{tabular}{ccp{4.1cm}}
		\hline
		Hyperparameter  &Value     \\ \hline
		Optimizer & Adam \cite{kingma2017adam}   \\ 
		Learning Rate   & 0.001     \\ 
		Batch Size   & 108       \\ 
		Replay buffer size  & 12288   \\ 
		Mapper update interval & 5 \\
		Mapper update batches & 20 \\
		Map scale & 0.02m \\
		Egocentric map size &  261 \\
		Global map size & 2001 \\
		Aggregation factor & 0.9 \\
		\hline
	\end{tabular}

	\label{tablehyper}
	
\end{table}


\addtolength{\textheight}{-2cm} 





\bibliographystyle{IEEEtran}

\bibliography{ref}

\begin{thebibliography}{10}
\providecommand{\url}[1]{#1}
\csname url@samestyle\endcsname
\providecommand{\newblock}{\relax}
\providecommand{\bibinfo}[2]{#2}
\providecommand{\BIBentrySTDinterwordspacing}{\spaceskip=0pt\relax}
\providecommand{\BIBentryALTinterwordstretchfactor}{4}
\providecommand{\BIBentryALTinterwordspacing}{\spaceskip=\fontdimen2\font plus
\BIBentryALTinterwordstretchfactor\fontdimen3\font minus
  \fontdimen4\font\relax}
\providecommand{\BIBforeignlanguage}[2]{{%
\expandafter\ifx\csname l@#1\endcsname\relax
\typeout{** WARNING: IEEEtran.bst: No hyphenation pattern has been}%
\typeout{** loaded for the language `#1'. Using the pattern for}%
\typeout{** the default language instead.}%
\else
\language=\csname l@#1\endcsname
\fi
#2}}
\providecommand{\BIBdecl}{\relax}
\BIBdecl

\bibitem{sep-aristotle-psychology}
C.~Shields, ``{Aristotle’s Psychology},'' in \emph{The {Stanford}
  Encyclopedia of Philosophy}, winter 2020~ed., E.~N. Zalta, Ed.\hskip 1em plus
  0.5em minus 0.4em\relax Metaphysics Research Lab, Stanford University, 2020.

\bibitem{weber_imagination-augmented_2018}
\BIBentryALTinterwordspacing
T.~Weber, S.~Racanière, D.~P. Reichert, L.~Buesing, A.~Guez, D.~J. Rezende,
  A.~P. Badia, O.~Vinyals, N.~Heess, Y.~Li, R.~Pascanu, P.~Battaglia,
  D.~Hassabis, D.~Silver, and D.~Wierstra, ``Imagination-{Augmented} {Agents}
  for {Deep} {Reinforcement} {Learning},'' \emph{arXiv:1707.06203 [cs, stat]},
  Feb. 2018, arXiv: 1707.06203. [Online]. Available:
  \url{http://arxiv.org/abs/1707.06203}
\BIBentrySTDinterwordspacing

\bibitem{hafner_dream_2020}
\BIBentryALTinterwordspacing
D.~Hafner, T.~Lillicrap, J.~Ba, and M.~Norouzi, ``Dream to {Control}:
  {Learning} {Behaviors} by {Latent} {Imagination},'' \emph{arXiv:1912.01603
  [cs]}, Mar. 2020, arXiv: 1912.01603. [Online]. Available:
  \url{http://arxiv.org/abs/1912.01603}
\BIBentrySTDinterwordspacing

\bibitem{churamani_clifer_2020}
N.~Churamani and H.~Gunes, ``{CLIFER}: {Continual} {Learning} with
  {Imagination} for {Facial} {Expression} {Recognition},'' May 2020.

\bibitem{ramakrishnan_occupancy_2020}
\BIBentryALTinterwordspacing
S.~K. Ramakrishnan, Z.~Al-Halah, and K.~Grauman, ``Occupancy {Anticipation} for
  {Efficient} {Exploration} and {Navigation},'' \emph{arXiv:2008.09285 [cs]},
  Aug. 2020, arXiv: 2008.09285. [Online]. Available:
  \url{http://arxiv.org/abs/2008.09285}
\BIBentrySTDinterwordspacing

\bibitem{mohajerin_multi_step_2019}
\BIBentryALTinterwordspacing
N.~Mohajerin and M.~Rohani, ``Multi-{Step} {Prediction} of {Occupancy} {Grid}
  {Maps} with {Recurrent} {Neural} {Networks},'' \emph{arXiv:1812.09395 [cs,
  stat]}, Jan. 2019, arXiv: 1812.09395. [Online]. Available:
  \url{http://arxiv.org/abs/1812.09395}
\BIBentrySTDinterwordspacing

\bibitem{ralph_neural_2017}
\BIBentryALTinterwordspacing
M.~A.~L. Ralph, E.~Jefferies, K.~Patterson, and T.~T. Rogers,
  ``\BIBforeignlanguage{en}{The neural and computational bases of semantic
  cognition},'' \emph{\BIBforeignlanguage{en}{Nature Reviews Neuroscience}},
  vol.~18, no.~1, pp. 42--55, Jan. 2017, number: 1 Publisher: Nature Publishing
  Group. [Online]. Available:
  \url{https://www.nature.com/articles/nrn.2016.150}
\BIBentrySTDinterwordspacing

\bibitem{mccormac_semanticfusion_2016}
\BIBentryALTinterwordspacing
J.~McCormac, A.~Handa, A.~Davison, and S.~Leutenegger, ``{SemanticFusion}:
  {Dense} {3D} {Semantic} {Mapping} with {Convolutional} {Neural} {Networks},''
  \emph{arXiv:1609.05130 [cs]}, Sep. 2016, arXiv: 1609.05130. [Online].
  Available: \url{http://arxiv.org/abs/1609.05130}
\BIBentrySTDinterwordspacing

\bibitem{chaplot_object_2020}
\BIBentryALTinterwordspacing
D.~S. Chaplot, D.~Gandhi, A.~Gupta, and R.~Salakhutdinov, ``Object {Goal}
  {Navigation} using {Goal}-{Oriented} {Semantic} {Exploration},''
  \emph{arXiv:2007.00643 [cs]}, Jul. 2020, arXiv: 2007.00643. [Online].
  Available: \url{http://arxiv.org/abs/2007.00643}
\BIBentrySTDinterwordspacing

\bibitem{Object_Semantic_Grid_Mapping2020}
Q.~Xianyu, W.~Wang, L.~Ziwei, X.~Zhang, D.~Yang, and R.~Wei, ``Object semantic
  grid mapping with 2d lidar and rgb-d camera for domestic robot navigation,''
  \emph{Applied Sciences}, vol.~10, p. 5782, 08 2020.

\bibitem{habitat19iccv}
M.~Savva, A.~Kadian, O.~Maksymets, Y.~Zhao, E.~Wijmans, B.~Jain, J.~Straub,
  J.~Liu, V.~Koltun, J.~Malik, D.~Parikh, and D.~Batra, ``Habitat: {A}
  {P}latform for {E}mbodied {AI} {R}esearch,'' in \emph{Proceedings of the
  IEEE/CVF International Conference on Computer Vision (ICCV)}, 2019.

\bibitem{daniel_theta_2010}
\BIBentryALTinterwordspacing
K.~Daniel, A.~Nash, S.~Koenig, and A.~Felner, ``\BIBforeignlanguage{en}{Theta*:
  {Any}-{Angle} {Path} {Planning} on {Grids}},''
  \emph{\BIBforeignlanguage{en}{Journal of Artificial Intelligence Research}},
  vol.~39, pp. 533--579, Oct. 2010. [Online]. Available:
  \url{https://jair.org/index.php/jair/article/view/10676}
\BIBentrySTDinterwordspacing

\bibitem{Lavalle98rapidly_exploringrandom}
S.~M. Lavalle, ``Rapidly-exploring random trees: A new tool for path
  planning,'' Tech. Rep., 1998.

\bibitem{Semantic_Mapping_Spatial_Concepts_2019}
Y.~Katsumata, A.~Taniguchi, Y.~Hagiwara, and T.~Taniguchi, ``Semantic mapping
  based on spatial concepts for grounding words related to places in daily
  environments,'' \emph{Frontiers in Robotics and AI}, vol.~6, 05 2019.

\bibitem{Matterport3D}
A.~Chang, A.~Dai, T.~Funkhouser, M.~Halber, M.~Niessner, M.~Savva, S.~Song,
  A.~Zeng, and Y.~Zhang, ``Matterport3d: Learning from rgb-d data in indoor
  environments,'' \emph{International Conference on 3D Vision (3DV)}, 2017.

\bibitem{he_mask_2018}
\BIBentryALTinterwordspacing
K.~He, G.~Gkioxari, P.~Dollár, and R.~Girshick, ``Mask {R}-{CNN},''
  \emph{arXiv:1703.06870 [cs]}, Jan. 2018, arXiv: 1703.06870. [Online].
  Available: \url{http://arxiv.org/abs/1703.06870}
\BIBentrySTDinterwordspacing

\bibitem{MS-coco}
T.-Y. Lin, M.~Maire, S.~Belongie, J.~Hays, P.~Perona, D.~Ramanan,
  P.~Doll{\'a}r, and C.~L. Zitnick, ``Microsoft coco: Common objects in
  context,'' in \emph{Computer Vision -- ECCV 2014}, D.~Fleet, T.~Pajdla,
  B.~Schiele, and T.~Tuytelaars, Eds.\hskip 1em plus 0.5em minus 0.4em\relax
  Cham: Springer International Publishing, 2014, pp. 740--755.

\bibitem{song2016semantic}
S.~Song, F.~Yu, A.~Zeng, A.~X. Chang, M.~Savva, and T.~Funkhouser, ``Semantic
  scene completion from a single depth image,'' 2016.

\bibitem{chaplot2020learning}
D.~S. Chaplot, D.~Gandhi, S.~Gupta, A.~Gupta, and R.~Salakhutdinov, ``Learning
  to explore using active neural slam,'' in \emph{International Conference on
  Learning Representations (ICLR)}, 2020.

\bibitem{he2015deep}
K.~{He}, X.~{Zhang}, S.~{Ren}, and J.~{Sun}, ``Deep residual learning for image
  recognition,'' in \emph{2016 IEEE Conference on Computer Vision and Pattern
  Recognition (CVPR)}, 2016, pp. 770--778.

\bibitem{ronneberger2015unet}
O.~Ronneberger, P.~Fischer, and T.~Brox, ``U-net: Convolutional networks for
  biomedical image segmentation,'' in \emph{Medical Image Computing and
  Computer-Assisted Intervention -- MICCAI 2015}, N.~Navab, J.~Hornegger, W.~M.
  Wells, and A.~F. Frangi, Eds.\hskip 1em plus 0.5em minus 0.4em\relax Cham:
  Springer International Publishing, 2015, pp. 234--241.

\bibitem{WangSCJDZLMTWLX19}
J.~Wang, K.~Sun, T.~Cheng, B.~Jiang, C.~Deng, Y.~Zhao, D.~Liu, Y.~Mu, M.~Tan,
  X.~Wang, W.~Liu, and B.~Xiao, ``Deep high-resolution representation learning
  for visual recognition,'' \emph{TPAMI}, 2019.

\bibitem{kingma2017adam}
D.~Kingma and J.~Ba, ``Adam: A method for stochastic optimization,''
  \emph{International Conference on Learning Representations}, 12 2014.

\end{thebibliography}

\end{document}